\documentclass[11pt]{article}

\usepackage[preprint]{acl}
\usepackage{times}
\usepackage{latexsym}
\usepackage[T1]{fontenc}
\usepackage[utf8]{inputenc}
\usepackage{microtype}
\usepackage{inconsolata}
\pdfmapfile{+zi4.map}
\usepackage{graphicx}
\usepackage{subcaption}
\usepackage{booktabs}
\usepackage{amsmath}
\usepackage{amssymb}
\usepackage{amsthm}
\usepackage{bm}
\usepackage{xcolor}
\usepackage{colortbl}
\usepackage{array}
\usepackage{multirow}

\definecolor{lightcyan1}{RGB}{240,250,252}
\definecolor{lightcyan2}{RGB}{215,238,244}
\definecolor{lightcyan3}{RGB}{182,222,233}
\definecolor{lightcyan4}{RGB}{140,202,217}
\definecolor{lightcyan5}{RGB}{95,178,196}
\definecolor{lightred1}{RGB}{253,242,242}
\definecolor{lightred2}{RGB}{250,222,222}
\definecolor{lightred3}{RGB}{246,196,196}
\definecolor{lightred4}{RGB}{238,162,162}
\definecolor{lightred5}{RGB}{226,120,120}

\usepackage{textcomp}
\usepackage{enumitem}
\usepackage{algorithm}
\usepackage{algpseudocode}

\newtheorem{lemma}{Lemma}

\graphicspath{{figures/}}

\title{A Token-Level Analysis of Sampled-Token Reverse-KL On-Policy Distillation}
\author{
  \textbf{Bing Shao\textsuperscript{1}}%
  \thanks{Equal contribution. $^\dagger$Corresponding author.} \quad
  \textbf{Jiazheng Zhang\textsuperscript{1}}\footnotemark[1] \quad
  \textbf{Long Ma\textsuperscript{1}} \quad
  \textbf{Yujiong Shen\textsuperscript{1}} \quad
  \textbf{Senjie Jin\textsuperscript{1}} \\
  \textbf{Xin Guo\textsuperscript{1}} \quad
  \textbf{Yuming Yang\textsuperscript{1}} \quad
  \textbf{Mingxu Chai\textsuperscript{1}} \quad
  \textbf{Zhiheng Xi\textsuperscript{1}} \quad
  \textbf{Boyang Liu\textsuperscript{1}} \\
  \textbf{Junlin Shang\textsuperscript{1}} \quad
  \textbf{Peixin Wang\textsuperscript{1}} \quad
  \textbf{Qi Zhang\textsuperscript{1}} \quad
  \textbf{Tao Gui\textsuperscript{1}}\textsuperscript{$\dagger$} \quad
  \textbf{Qi Zhang\textsuperscript{1}} \quad
  \textbf{Xuanjing Huang\textsuperscript{1}} \\[0.5em]
  \textsuperscript{1}College of Computer Science and Artificial Intelligence,
  Fudan University \\
  \texttt{bshao25@m.fudan.edu.cn}, \quad \texttt{tgui@fudan.edu.cn}
}

\hypersetup{
  pdftitle={A Token-Level Analysis of Sampled-Token Reverse-KL On-Policy Distillation},
  pdfauthor={Bing Shao, Jiazheng Zhang, Long Ma, Yujiong Shen, Senjie Jin, Xin Guo, Yuming Yang, Mingxu Chai, Zhiheng Xi, Boyang Liu, Junlin Shang, Peixin Wang, Qi Zhang, Tao Gui, Qi Zhang, Xuanjing Huang}
}


\begin{document}
\maketitle

\begin{abstract}
On-policy distillation (OPD) supervises a student on its own
trajectories with token-level signals from a frozen teacher, yet how a
sampled loss allocates updates across tokens remains poorly understood.
We analyze the gradient of the per-token K2 estimator of reverse KL
with respect to the student logits. The $\ell_1$ norm of this gradient
factorizes into the absolute teacher--student log-probability gap and a student-side softmax
factor that grows as the sampled token becomes less likely under the student.
In our math-distillation runs, these per-token norms are highly non-uniform:
low-student-probability tokens account for a disproportionate share of their
sum and are also enriched in large teacher--student gaps. As a
lightweight intervention suggested by this analysis, we study
Surprise-aware Reweighting (SuRe), a detached, bounded weighting rule that further
amplifies this existing allocation. Across two Qwen3 student scales,
SuRe improves several math metrics over vanilla OPD and shows no clear
degradation on the selected out-of-domain benchmarks. Our primary contribution is therefore a gradient-level characterization of reverse-KL OPD trained with the K2 estimator, with SuRe as one empirical instantiation.
\end{abstract}

\begin{figure*}[t]
  \centering
  \includegraphics[width=\textwidth]{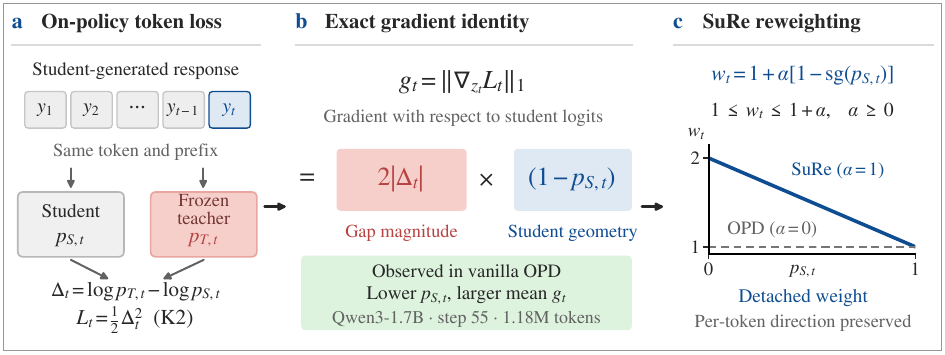}
  \caption{\textbf{From token-level gradient analysis to SuRe.}
  (a) The same student-sampled token and prefix are scored by the student
  and frozen teacher.
  (b) Exact K2 student-logit gradient norm for a fixed sample;
  the conditional-mean diagnostic is observational.
  (c) SuRe applies detached, bounded weights that preserve per-token
  gradient direction. Curves show analytic weight functions.}
  \label{fig:overview}
\end{figure*}

\section{Introduction}
\label{sec:intro}

Open language model families increasingly share a multi-stage
post-training recipe: supervised fine-tuning (SFT) initializes a general
policy, reinforcement learning (RL) develops one or more specialized
policies, and a final on-policy distillation stage either transfers
capabilities from a large teacher to a lightweight student or consolidates
capabilities from multiple policies into a unified model. Qwen3
\citep{DBLP:journals/corr/abs-2505-09388}, MiMo
\citep{DBLP:journals/corr/abs-2601-02780}, DeepSeek-V4
\citep{deepseekv4}, Kimi K3
\citep{DBLP:journals/corr/abs-2607-24653}, and GLM-5
\citep{DBLP:journals/corr/abs-2602-15763} are representative examples of
this broader pipeline.
\emph{On-policy distillation} (OPD) keeps training on-policy to reduce distribution shift while providing dense token-level feedback \citep{lu2025onpolicydistillation,DBLP:conf/iclr/Gu0WH24,DBLP:conf/iclr/AgarwalVZSGGB24,DBLP:journals/corr/abs-2602-12125}.
Throughout this paper, we study the gradient of the per-token K2 estimator
of reverse KL with respect to the student logits. At the level of the
immediate per-token distillation signal, the gradient-descent update induced
by K2 has the same direction as the unclipped policy-gradient update induced
by Kimi K3's detached sampled-token log-ratio OPD reward; Kimi K3 clips that
reward for stability
\citep[Sec.~4.1.3]{DBLP:journals/corr/abs-2607-24653}.
Understanding \emph{which} tokens receive large gradients, and \emph{why}
they receive them, is the question we take up in this paper.

Many token-level studies use quantities such as entropy or sampled-token
probability to analyze, select, or reweight token positions
\citep{DBLP:journals/corr/abs-2506-01939,DBLP:journals/corr/abs-2603-07079,DBLP:journals/corr/abs-2603-22117,DBLP:journals/corr/abs-2603-11137,li2026rethinking}.
These quantities provide useful views of token uncertainty and disagreement.
Our focus is complementary: we directly characterize the realized per-token
gradient of the sampled OPD loss with respect to the student logits and study
how the resulting gradient norms are allocated across sampled trajectories.

For the K2 estimator, a closed-form expansion of this gradient shows that its
$\ell_1$
norm factorizes into an absolute teacher--student gap and the student-side
term $1-\pi_S(y_t\,|\,c_t)$. In our runs, low-student-probability
positions account for a disproportionate share of the summed per-token
norms, and large teacher--student gaps are enriched in the same region. Thus the observed
concentration is a joint empirical pattern, rather than a consequence of
the student-side factor alone.

To test whether this allocation can guide optimization, we use
\textbf{SuRe} (\emph{Surprise-aware Reweighting}) as a lightweight
analysis-inspired intervention. It attaches a detached, bounded per-token
weight that mildly up-weights surprise tokens and approaches one for confident
tokens, with a single coefficient $\alpha$ that recovers vanilla OPD at
$\alpha=0$. SuRe requires no additional reference model, extra forward pass,
learned selector, or hard threshold. Distilling Qwen3-8B into
Qwen3-1.7B-Base and Qwen3-4B-Base on DeepMath, SuRe improves vanilla OPD
by up to $+6.7$pp on AIME24 and $+7.5$pp on AMC23 (pass@$8$) on the 1.7B
student and improves several metrics on the 4B student, while showing no
clear degradation on the selected OOD benchmarks.

\paragraph{Contributions.}
\begin{itemize}
\item \textbf{Gradient identity for the K2 estimator.}
We derive its gradient with respect to the student logits and
show that its $\ell_1$ norm factorizes into the absolute
teacher--student gap and $1-\pi_S(y_t\,|\,c_t)$.

\item \textbf{Empirical token-level allocation.}
In the evaluated Qwen3 math setup, low-student-probability tokens account
for a disproportionate share of the sum of these gradient norms and are
enriched in large teacher--student gaps.

\item \textbf{An analysis-inspired intervention.}
We study SuRe as one bounded amplification rule. It improves several
metrics, while the controls do not fully separate exact surprise
assignment from a broader benefit of non-uniform weighting.
\end{itemize}

\section{Preliminaries}
\label{sec:prelim}

\subsection{On-policy Distillation}
\label{sec:prelim-opd}

Unlike OPD formulations that place KL in a reward signal and optimize it
through policy gradient \citep{lu2025onpolicydistillation}, we study a
loss-based formulation in which the per-token K2 estimator is backpropagated
directly, with no separate policy-gradient term, following prior loss-based
work \citep{DBLP:conf/iclr/AgarwalVZSGGB24}. Student rollouts are treated as
fixed samples within each update.

\paragraph{Notation.}
Let $\pi_T$ be a frozen \emph{teacher} model and $\pi_\theta$ the \emph{student}
model being trained.
For a problem prompt $x$, the student generates a response
$y=(y_1,\dots,y_L)$ token by token.
We write $c_t\triangleq(x,y_{<t})$ for the context at decoding position $t$ and write $\pi_S$
for the student policy $\pi_\theta$ when no ambiguity arises.
Let $\mathcal{V}$ be the vocabulary; the student's next-token distribution at
position $t$ is
\begin{equation}
  \pi_S(v\,|\,c_t)
    = \frac{\exp(z_v)}{\sum_{u\in\mathcal{V}}\exp(z_u)},
  \ v\in\mathcal{V},
  \label{eq:prelim-softmax}
\end{equation}
where $z\in\mathbb{R}^{|\mathcal{V}|}$ are the student logits at position $t$.
Training prompts are drawn from a fixed prompt set
$\mathcal{D}_x \triangleq \{x^{(i)}\}_{i=1}^{N}$, and student responses
$y\sim\pi_\theta(\cdot\,|\,x)$ are sampled on-policy.

\paragraph{Kullback--Leibler divergence.}
For two distributions $p,q$ over $\mathcal{V}$,
\begin{equation}
  D_{\mathrm{KL}}(p\,\|\,q) = \sum_{v\in\mathcal{V}}p(v)\log\frac{p(v)}{q(v)}.
  \label{eq:prelim-kl}
\end{equation}
We instantiate it on next-token distributions and use the \emph{reverse}
KL $D_{\mathrm{KL}}(\pi_S(\cdot\,|\,c_t)\,\|\,\pi_T(\cdot\,|\,c_t))$ as the
distillation objective.

\paragraph{Sampled token-level reverse-KL objective.}
We optimize the K2 estimator of reverse KL at each token on student
rollouts.
For each sampled token $y_t$, define the teacher--student log-probability gap
\begin{equation}
  \Delta\log p_t
    \triangleq \log\pi_T(y_t\,|\,c_t) - \log\pi_S(y_t\,|\,c_t).
  \label{eq:prelim-dlp}
\end{equation}
We use the K2 estimator \citep{schulman-kl} at each token,
\begin{equation}
  L_t^{\mathrm{RKL}} \triangleq \tfrac{1}{2}\bigl(\Delta\log p_t\bigr)^2,
  \label{eq:prelim-Lt}
\end{equation}
and aggregate over valid response tokens with a token-mean denominator,
\begin{equation}
  \begin{aligned}
    \mathcal{L}_{\mathrm{RKL}}
    \triangleq{}& \mathbb{E}_{x\sim\mathcal{D}_x,\,y\sim\pi_\theta(\cdot\,|\,x)} \\
    &\left[\frac{1}{N_{\mathrm{valid}}}\sum_t m_t\,L_t^{\mathrm{RKL}}\right],
  \end{aligned}
  \label{eq:prelim-rkl}
\end{equation}
where $m_t\in\{0,1\}$ is the response mask and
$N_{\mathrm{valid}}=\sum_t m_t$.
Equation~\eqref{eq:prelim-Lt} is evaluated on one token sampled from the
current student distribution. Although K2 is a biased estimator of the
reverse-KL value, its realized-loss logit gradient is unbiased in expectation
under current-student sampling at a fixed context $c_t$:
\begin{equation}
\begin{aligned}
&\mathbb{E}_{y_t\sim\pi_S(\cdot\,|\,c_t)}
\!\left[\nabla_z L_t^{\mathrm{RKL}}\right] \\
&\quad= \nabla_z D_{\mathrm{KL}}\!\left(
\pi_S(\cdot\,|\,c_t)\,\|\,\pi_T(\cdot\,|\,c_t)
\right).
\end{aligned}
\label{eq:prelim-k2-gradient-expectation}
\end{equation}
Here $y_t$ is sampled from the current student and then held fixed during
backpropagation. This identity does not differentiate through the
sampled trajectory or the distribution of prefixes. The training objective is
$\mathcal{L}_{\mathrm{RKL}}$ alone, with no separate policy-gradient term, and
the teacher outputs are treated as fixed.

\paragraph{Softmax gradient identity.}
The analysis in Sec.~\ref{sec:analysis} works with the gradient of
$L_t^{\mathrm{RKL}}$ with respect to the position-$t$ logit vector
$z\in\mathbb{R}^{|\mathcal{V}|}$ defined in \eqref{eq:prelim-softmax}.
Let $e_{y_t}\in\mathbb{R}^{|\mathcal{V}|}$ be the one-hot indicator for the
sampled token $y_t$. The standard softmax Jacobian gives
\begin{equation}
  \nabla_z\log\pi_S(y_t\,|\,c_t)
    = e_{y_t} - \pi_S(\cdot\,|\,c_t).
  \label{eq:prelim-grad-logps}
\end{equation}
The analysis in Sec.~\ref{sec:analysis-identity} chains this
identity with $\nabla_{\log\pi_S(y_t\,|\,c_t)}L_t^{\mathrm{RKL}}=-\Delta\log p_t$
to obtain the closed form for $\nabla_z L_t^{\mathrm{RKL}}$.

\subsection{Diagnostic Quantities}
\label{sec:prelim-diag}

We also use entropy and Jensen--Shannon divergence as diagnostics on
next-token distributions.

\paragraph{Entropy.} For a distribution $p$ over $\mathcal{V}$, the
Shannon entropy is
\begin{equation}
  H(p) = -\sum_{v\in\mathcal{V}}p(v)\log p(v),
  \label{eq:prelim-entropy-def}
\end{equation}
which quantifies the uncertainty of $p$.

\paragraph{Jensen--Shannon divergence.}
With $M\triangleq\tfrac{1}{2}(p+q)$,
\begin{equation}
  \mathrm{JSD}(p,q) = \tfrac{1}{2}D_{\mathrm{KL}}(p\,\|\,M) + \tfrac{1}{2}D_{\mathrm{KL}}(q\,\|\,M),
  \label{eq:prelim-jsd-def}
\end{equation}
which lies in $[0,\log 2]$. At position $t$ we write
\begin{equation}
  \mathrm{JSD}_t \triangleq \mathrm{JSD}\!\bigl(\pi_S(\cdot\,|\,c_t),\pi_T(\cdot\,|\,c_t)\bigr).
  \label{eq:prelim-jsd}
\end{equation}
Unlike the signed sampled-token gap $\Delta\log p_t$, $\mathrm{JSD}_t$
summarizes full distributions and discards direction.

\section{Token-Level Gradient Analysis}
\label{sec:analysis}
We analyze vanilla reverse-KL OPD with a Qwen3-1.7B-Base student,
Qwen3-8B teacher, and DeepMath-hard (Sec.~\ref{sec:experiments-setup}).
First, we compare token probabilities at initialization and after training.
We then analyze the K2 estimator's student-logit gradient on OPD rollouts
and a separate mid-training dump. The endpoint checkpoint shift is distinct
from the teacher--student residual used in the gradient analysis.

\subsection{Post-training checkpoint-shift diagnostic}
\label{sec:analysis-traj}

Here, \emph{Base} denotes the untouched Qwen3-1.7B-Base initialization,
whereas \emph{OPD} denotes the final vanilla-OPD checkpoint at step
$222$. Each checkpoint independently generates responses on the same
DeepMath-hard mixture using temperature $1.0$, top-$p$ $1.0$, and seed
$42$; we call the resulting sets Base rollouts and OPD rollouts. For every
realized token $y_t$ and prefix $c_t$ in either set, we hold both fixed and
score the token under the two frozen checkpoints. The pooled analysis
covers $1.84$M response tokens. We define the signed endpoint shift as
\begin{equation}
\begin{aligned}
\Delta_{\mathrm{OPD-Base},t}
  &= \log \pi_{\mathrm{OPD}}(y_t\,|\,c_t) \\
  &{}- \log \pi_{\mathrm{Base}}(y_t\,|\,c_t),
\end{aligned}
\label{eq:analysis-checkpoint-shift}
\end{equation}
Positive values indicate higher sampled-token probability under OPD.
This statistic has no teacher term, unlike the K2 residual $\Delta\log p_t$.

\begin{figure*}[t]
  \centering
  \begin{subfigure}[b]{0.49\linewidth}
    \centering
    \includegraphics[width=\linewidth]{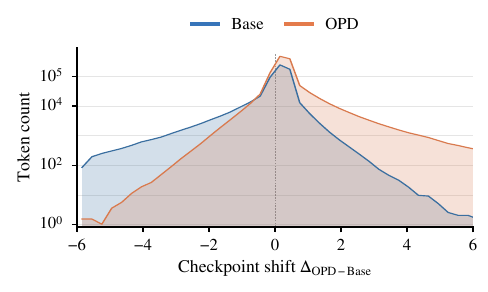}
    \caption{Long-tail overlay (log-count).}
    \label{fig:trajectory-diagnostic-longtail}
  \end{subfigure}\hfill
  \begin{subfigure}[b]{0.49\linewidth}
    \centering
    \includegraphics[width=\linewidth]{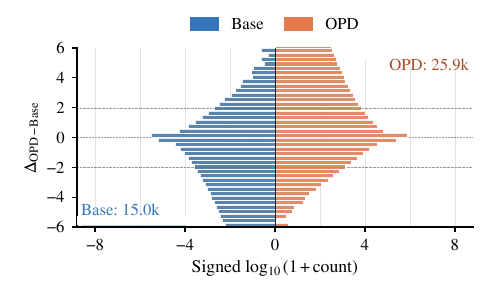}
    \caption{Tornado view (signed log-count).}
    \label{fig:trajectory-diagnostic-tornado}
  \end{subfigure}
  \caption{\textbf{Signed post-training checkpoint shift.}
  Curves are grouped by the checkpoint that generated each rollout; every
  realized token and prefix is rescored under both checkpoints. Base
  rollouts lean negative, whereas OPD rollouts exhibit a positive tail.
  Panels show a log-count overlay (a) and a signed tornado view (b);
  annotations in (b) count tokens with $|\Delta_{\mathrm{OPD-Base}}|\geq2$.}
  \label{fig:trajectory-diagnostic}
\end{figure*}

\paragraph{Checkpoint shift.}
Only $8.5\%$ of OPD-rollout tokens and $7.1\%$ of Base-rollout tokens
satisfy $|\Delta_{\mathrm{OPD-Base}}|>1$
(Figure~\ref{fig:trajectory-diagnostic}). Base rollouts lean negative:
OPD assigns lower probability to many Base-sampled tokens. OPD rollouts
instead exhibit a positive tail. This describes endpoint model change,
not teacher endorsement or gradient concentration; we examine the
loss-level quantities next.

\subsection{An exact identity for the gradient norm of the K2 estimator}
\label{sec:analysis-identity}

Here $\Delta\log p_t$ compares the frozen Qwen3-8B teacher with the
student at the same sampled token and prefix. Holding the sample and
teacher output fixed, the chain rule applied to
Eqs.~\eqref{eq:prelim-Lt} and~\eqref{eq:prelim-grad-logps} gives
\begin{equation}
\nabla_z L_t^{\mathrm{RKL}}
  = -\Delta\log p_t\bigl(e_{y_t}-\pi_S(\cdot\,|\,c_t)\bigr).
\label{eq:grad-vector}
\end{equation}
Taking the $\ell_1$ norm gives the diagnostic identity used below.

\begin{lemma}[Gradient norm of the per-token K2 estimator]
\label{lem:opd-grad-norm}
For the per-token estimator in \eqref{eq:prelim-Lt},
\begin{equation}
\begin{aligned}
\bigl\|\nabla_z L_t^{\mathrm{RKL}}\bigr\|_1
  &= 2\,\bigl|\Delta\log p_t\bigr| \\
  &{}\cdot \bigl(1-\pi_S(y_t\,|\,c_t)\bigr).
\end{aligned}
\label{eq:opd-grad-l1}
\end{equation}
\end{lemma}

The proof, including the justification for using the $\ell_1$ norm, is
given in Appendix~\ref{app:gradient}.

\paragraph{Direction and magnitude.}
The sign of $\Delta\log p_t$ determines whether the sampled token is raised
or suppressed, while the magnitude factorizes into a teacher--student gap
and the student-side geometry term $(1-\pi_S)$. Thus, in OPD, the
low-probability effect should be interpreted on the student side: holding
$|\Delta\log p_t|$ fixed, tokens assigned smaller
$\pi_S(y_t\,|\,c_t)$ receive larger gradient coefficients.

\begin{figure*}[t]
  \centering
  \includegraphics[width=\linewidth]{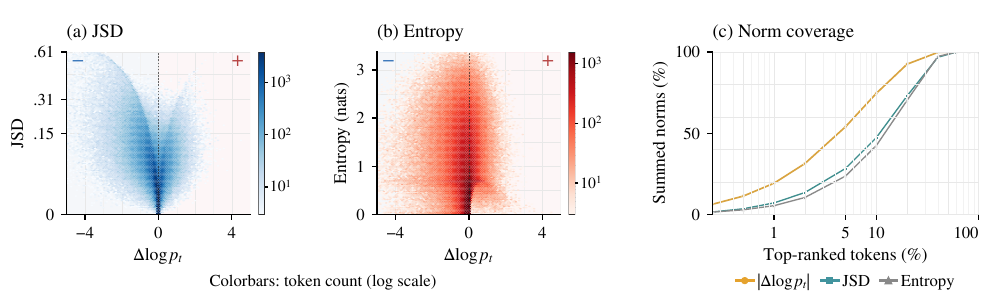}
  \caption{\textbf{Why signed residuals rather than entropy/JSD.}
  Panels (a,b) compare the teacher--OPD residual gap with JSD and OPD
  entropy; most tokens sit in a low-gap, low-uncertainty bulk. Panel
  (c) ranks tokens by each statistic and measures the share of the sum of
  gradient norms covered by each ranking. JSD and entropy
  use a top-50 approximation, computed
  by restricting each position's full-distribution diagnostic to its
  50 highest-probability candidate tokens.}
  \label{fig:why-dlp}
\end{figure*}

\paragraph{Diagnostic rationale.}
On rollouts from the final OPD checkpoint, entropy and JSD are
non-directional and mostly small. Ranking by
$|\Delta\log p_t|$ captures a much larger share of the sum of these
gradient norms:
$54.1\%/74.6\%$ in the top $5\%/10\%$ tokens, versus $28.4\%/47.3\%$ for
JSD and $23.7\%/42.4\%$ for entropy (Fig.~\ref{fig:why-dlp}). The signed
residual retains update direction, and its absolute value is the stronger
scalar ranker in this comparison. The next subsection examines how this
norm varies with student probability.

\subsection{Student probability and concentration of gradient norms}
\label{sec:analysis-ssc}

\begin{figure*}[t]
  \centering
  \includegraphics[width=\linewidth]{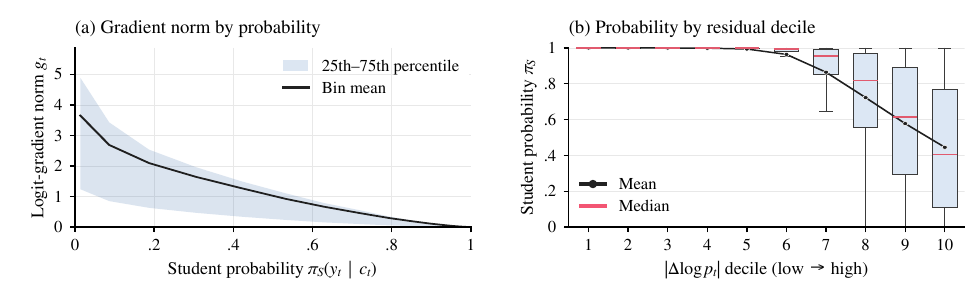}
  \caption{\textbf{Mean logit-gradient norms are larger at
  low-student-probability positions.} Panel (a) shows the mean per-token
  logit-gradient $\ell_1$ norm within student-probability bins, with
  shading for the 25th--75th percentiles. Panel (b) shows student
  probability distributions within absolute-residual deciles; the
  connected points indicate mean student probability.}
  \label{fig:grad-vs-pis}
\end{figure*}

At fixed $|\Delta\log p_t|$, Eq.~\eqref{eq:opd-grad-l1} scales with
$1-\pi_S(y_t\,|\,c_t)$. We test this on a separate $1.18$M-token dump
from step $55$ of vanilla OPD, rather than either endpoint checkpoint.
We compute
\begin{equation}
g_t = 2\,|\Delta\log p_t|\,\bigl(1-\pi_S(y_t\,|\,c_t)\bigr),
\label{eq:grad-coeff}
\end{equation}
and sum $g_t$ within $\pi_S$ deciles
(Figure~\ref{fig:grad-vs-pis}).

\paragraph{Observed concentration.}
In this snapshot, lower-$\pi_S$ bins account for a larger share of the
sum of these gradient norms, and large absolute gaps are enriched in the
same bins (Figure~\ref{fig:grad-vs-pis}). This is an empirical association: the
factor $1-\pi_S$ alone does not imply the observed concentration. A
probability-only weight therefore amplifies all low-$\pi_S$ samples,
regardless of the sign of the teacher--student gap.

\section{An Analysis-Inspired Reweighting}
\label{sec:method}

\textbf{SuRe (Surprise-aware Reweighting).}
To probe whether the observed allocation is useful for optimization, we
study SuRe as a simple intervention. The gradient identity motivates
using student probability, but it does not uniquely imply the affine rule
below.

\begin{table*}[t!]
    \centering
    \caption{\textbf{In-domain performance (\%).}
    For AIME24, AIME25, and AMC23 we report avg@$8$ and pass@$8$;
    for MATH-500 we report avg@$4$ and pass@$4$.
    Cells with deeper background color correspond to better performance within each model group.}

    \label{tab:main}
    \small
    \setlength{\tabcolsep}{4.5pt}
    \begin{tabular}{l|cc|cc|cc|cc}
    \toprule[1.2pt]
        & \multicolumn{2}{c|}{\textbf{AIME24}}
        & \multicolumn{2}{c|}{\textbf{AIME25}}
        & \multicolumn{2}{c|}{\textbf{AMC23}}
        & \multicolumn{2}{c}{\textbf{MATH-500}} \\
        \cmidrule{2-3} \cmidrule{4-5} \cmidrule{6-7} \cmidrule{8-9}
        \textbf{Methods}
        & avg@$8$ & pass@$8$
        & avg@$8$ & pass@$8$
        & avg@$8$ & pass@$8$
        & avg@$4$ & pass@$4$ \\
        \midrule
            \multicolumn{9}{c}{\textit{\textbf{Qwen3-1.7B-Base}}} \\
        \midrule
        - Base
        & \cellcolor{lightcyan1}{4.17}  & \cellcolor{lightcyan1}{16.67}
        & \cellcolor{lightcyan1}{4.17}  & \cellcolor{lightcyan3}{16.67}
        & \cellcolor{lightcyan1}{30.00} & \cellcolor{lightcyan1}{65.00}
        & \cellcolor{lightcyan1}{56.35} & \cellcolor{lightcyan3}{76.80} \\

        - KD
        & \cellcolor{lightcyan1}{7.08}  & \cellcolor{lightcyan1}{20.00}
        & \cellcolor{lightcyan1}{3.33}  & \cellcolor{lightcyan1}{13.33}
        & \cellcolor{lightcyan1}{30.94} & \cellcolor{lightcyan1}{67.50}
        & \cellcolor{lightcyan1}{56.45} & \cellcolor{lightcyan1}{75.60} \\

        - SeqKD
        & \cellcolor{lightcyan1}{5.83}  & \cellcolor{lightcyan1}{20.00}
        & \cellcolor{lightcyan1}{4.17}  & \cellcolor{lightcyan3}{20.00}
        & \cellcolor{lightcyan1}{32.19} & \cellcolor{lightcyan3}{72.50}
        & \cellcolor{lightcyan1}{56.25} & \cellcolor{lightcyan1}{74.80} \\

        - Vanilla OPD
        & \cellcolor{lightcyan3}{9.17}  & \cellcolor{lightcyan1}{16.67}
        & \cellcolor{lightcyan3}{5.83}  & \cellcolor{lightcyan1}{16.67}
        & \cellcolor{lightcyan3}{39.38} & \cellcolor{lightcyan1}{67.50}
        & \cellcolor{lightcyan3}{66.55} & \cellcolor{lightcyan5}{\textbf{80.80}} \\

        - SuRe
        & \cellcolor{lightcyan5}{\textbf{9.58}}  & \cellcolor{lightcyan5}{\textbf{23.33}}
        & \cellcolor{lightcyan5}{\textbf{7.08}}  & \cellcolor{lightcyan1}{16.67}
        & \cellcolor{lightcyan5}{\textbf{43.12}} & \cellcolor{lightcyan5}{\textbf{75.00}}
        & \cellcolor{lightcyan5}{\textbf{67.65}} & \cellcolor{lightcyan5}{\textbf{80.80}} \\

        \specialrule{1pt}{0.4ex}{0.4ex}
            \multicolumn{9}{c}{\textit{\textbf{Qwen3-4B-Base}}} \\
        \midrule
        - Base
        & \cellcolor{lightred1}{12.08} & \cellcolor{lightred1}{26.67}
        & \cellcolor{lightred1}{7.08}  & \cellcolor{lightred1}{23.33}
        & \cellcolor{lightred1}{39.06} & \cellcolor{lightred1}{75.00}
        & \cellcolor{lightred1}{57.65} & \cellcolor{lightred1}{81.60} \\

        - KD
        & \cellcolor{lightred1}{10.00} & \cellcolor{lightred1}{20.00}
        & \cellcolor{lightred1}{11.67} & \cellcolor{lightred3}{33.33}
        & \cellcolor{lightred1}{32.50} & \cellcolor{lightred1}{75.00}
        & \cellcolor{lightred1}{47.35} & \cellcolor{lightred1}{76.00} \\

        - SeqKD
        & \cellcolor{lightred1}{8.33}  & \cellcolor{lightred1}{30.00}
        & \cellcolor{lightred1}{5.83}  & \cellcolor{lightred1}{20.00}
        & \cellcolor{lightred1}{32.81} & \cellcolor{lightred1}{70.00}
        & \cellcolor{lightred1}{55.40} & \cellcolor{lightred1}{82.60} \\

        - Vanilla OPD
        & \cellcolor{lightred3}{18.75} & \cellcolor{lightred1}{30.00}
        & \cellcolor{lightred5}{\textbf{17.08}} & \cellcolor{lightred5}{\textbf{40.00}}
        & \cellcolor{lightred1}{56.56} & \cellcolor{lightred1}{85.00}
        & \cellcolor{lightred1}{79.05} & \cellcolor{lightred3}{87.20} \\

        - SuRe
        & \cellcolor{lightred5}{\textbf{19.58}} & \cellcolor{lightred5}{\textbf{36.67}}
        & \cellcolor{lightred1}{14.17} & \cellcolor{lightred3}{36.67}
        & \cellcolor{lightred5}{\textbf{58.44}} & \cellcolor{lightred5}{\textbf{90.00}}
        & \cellcolor{lightred5}{\textbf{79.15}} & \cellcolor{lightred5}{\textbf{88.40}} \\

    \bottomrule[1.2pt]
    \end{tabular}

\end{table*}

\subsection{Why a Student-side Reweighting}
\label{sec:method-motivation}
The factor $1-\pi_S$ captures student softmax geometry rather than
teacher uncertainty. Unlike low-$\pi_S$ masking, reweighting retains
surprised tokens without hard thresholds. We test smooth, bounded
amplification, $w_t\in[1,1+\alpha]$, with one parameter recovering
vanilla OPD. The analysis does not determine whether this concentration
should be amplified, attenuated, normalized, or clipped.

\subsection{A Factor-Targeted Instantiation}
\label{sec:method-rule}
We make this concentration controllable through a multiplicative
weight on the per-token loss. Define the detached student probability
of the sampled token,
\begin{equation}
\bar{\pi}_{S,t}=\mathrm{sg}\bigl(\pi_S(y_t\,|\,c_t)\bigr),
\label{eq:method-detached}
\end{equation}
and the SuRe weight
\begin{equation}
w_t = 1 + \alpha\bigl(1-\bar{\pi}_{S,t}\bigr),\ \alpha\ge 0.
\label{eq:method-weight}
\end{equation}
Because $\bar{\pi}_{S,t}$ is detached, $w_t$ simply rescales the
baseline per-token gradient. Substituting it into equation \eqref{eq:grad-coeff}
yields
\begin{equation}
\begin{aligned}
w_t\,g_t
  &= 2\bigl|\Delta\log p_t\bigr| \\
  &{}\cdot
    \bigl[(1-\pi_S)+\alpha(1-\pi_S)^2\bigr].
\end{aligned}
\label{eq:grad-quadratic}
\end{equation}
SuRe preserves the gap factor and emphasizes lower-probability tokens.
Since $w_t\geq1$ and the denominator is not renormalized, it can also
increase the overall loss scale.

\subsection{The SuRe Objective}
\label{sec:method-wsum}

Let $\mathcal{T}$ denote the set of valid response tokens after the
response mask. We apply $w_t$ to the per-token reverse-KL loss in
\eqref{eq:prelim-rkl} while keeping the unweighted token-mean
denominator:
\begin{equation}
\mathcal{L}_{\mathrm{RKL}\text{-}\mathrm{rw}}
  = \frac{1}{|\mathcal{T}|}\sum_{t\in\mathcal{T}} w_t\,L_t^{\mathrm{RKL}}.
\label{eq:method-wsum}
\end{equation}

Appendix~\ref{app:sure-algorithm} gives the pseudocode, isolating the
only implementation change: the detached scalar $w_t$.

\section{Experiments}
\label{sec:experiments}

\subsection{Setup}
\label{sec:experiments-setup}

\paragraph{Models and data.}
We distill from Qwen3-8B \citep{DBLP:journals/corr/abs-2505-09388}
into Qwen3-1.7B-Base and Qwen3-4B-Base students.
Training uses the $57$K hard split (difficulty $\geq 6$) of
DeepMath \citep{deepmath}.

\paragraph{Training.}
The main experiments use seed $42$. The second-seed check uses seed $43$;
all other settings are held fixed, and the corresponding OPD and SuRe
checkpoints are evaluated at step $222$.
We train for two epochs on $32{\times}$H20 with learning rate
$10^{-6}$ and batch size $512$. Full configs are in
Appendix~\ref{app:exp}. The second seed is limited to the MATH-500 controls.

\paragraph{Evaluation.}
Math reasoning benchmarks include AIME2024 \citep{Zhang:2024:AIME},
AIME2025 \citep{Zhang:2025:AIME}, AMC23 \citep{Li:2024:AMC}, and
MATH-500 \citep{Lightman:2024:MATH500}.
Out-of-domain (OOD) benchmarks include code input/output prediction (CRUX, \citealp{Gu:2024:CRUX}),
instruction following (IFEval, \citealp{Zhou:2023:IFEval}), and general ability (MMLU-Pro, \citealp{Wang:2024:MMLU}).
For competition benchmarks (AIME24, AIME25, AMC23) we report
both avg@$8$ and pass@$8$ with temperature $0.7$ and top-$p$ $0.9$;
for MATH-500 we report avg@$4$ and pass@$4$; for OOD benchmarks
we report pass@$1$.

\subsection{Main Results}
\label{sec:experiments-main}

\paragraph{Math performance.}
At both scales, vanilla OPD exceeds Base, KD
\citep{DBLP:journals/corr/HintonVD15}, and SeqKD
\citep{DBLP:conf/emnlp/KimR16} in avg@$k$ on every math benchmark
(Table~\ref{tab:main}). KD and SeqKD sometimes underperform Base,
including on 4B AMC23; these experiments do not isolate why.

SuRe ($\alpha=1.0$) raises AMC23 avg@$8$/pass@$8$ by $3.7/7.5$pp at
1.7B and pass@$8$ by $5.0$pp at 4B. MATH-500 gains are small, while
both AIME25 metrics decrease at 4B, so the benefit is not uniform.
Appendix~\ref{app:grpo} compares with GRPO under different training
signals; it is not a matched comparison of equivalent objectives.

\begin{figure*}[t!]
    \centering
    \begin{subfigure}[t]{0.33\textwidth}
        \centering
        \includegraphics[width=\linewidth]{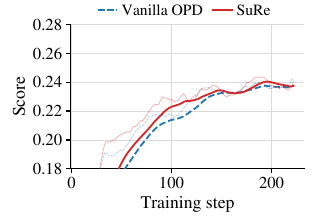}
        \caption{Mean rollout score}
        \label{fig:train-dyn-score}
    \end{subfigure}%
    \begin{subfigure}[t]{0.33\textwidth}
        \centering
        \includegraphics[width=\linewidth]{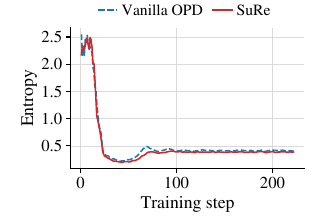}
        \caption{Actor entropy}
        \label{fig:train-dyn-ent}
    \end{subfigure}%
    \begin{subfigure}[t]{0.33\textwidth}
        \centering
        \includegraphics[width=\linewidth]{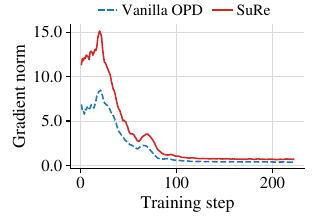}
        \caption{Gradient norm}
        \label{fig:train-dyn-grad}
    \end{subfigure}
    \caption{\textbf{Training dynamics on Qwen3-1.7B-Base.}
    Vanilla OPD (blue dashed) vs.\ SuRe at $\alpha{=}1.0$ (red).
    SuRe improves mean rollout score and increases early gradient norm
    while tracking OPD's actor entropy.}
    \label{fig:train-dynamics}

\end{figure*}

\begin{table}[t!]
    \centering
    \includegraphics[width=\linewidth]{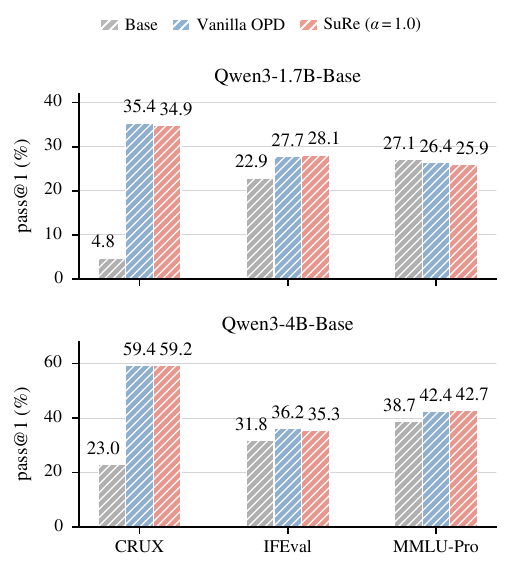}
    \captionof{figure}{\textbf{Out-of-domain performance (pass@$1$, \%).}
    Bars compare Base, Vanilla OPD, and SuRe on CRUX, IFEval, and
    MMLU-Pro for both student scales. On the evaluated OOD tasks,
    OPD and SuRe are broadly comparable, but neither uniformly improves
    over the base; both are slightly below the base on 1.7B MMLU-Pro.}
    \label{fig:ood}

\end{table}

\paragraph{Selected out-of-domain results are mixed.}
Although training data is restricted to hard math
(DeepMath, difficulty $\geq 6$), Figure~\ref{fig:ood} shows that OPD and
SuRe are broadly comparable on CRUX, IFEval, and MMLU-Pro. On 1.7B
MMLU-Pro, both are slightly below the base. We therefore observe neither
uniform OOD improvement nor clear evidence of a broad transfer effect on
these selected tasks.

\paragraph{Training dynamics show direct amplification.}
Figure~\ref{fig:train-dynamics} shows that SuRe increases the early
gradient norm while actor entropy remains similar. Since the unnormalized
weights have mean above one, the larger norm is expected and cannot by
itself distinguish surprise alignment from a larger effective update
scale.

\subsection{Ablations and Controls}
\label{sec:experiments-controls}

\begin{figure*}[t!]
    \centering
    \includegraphics[width=0.82\textwidth]{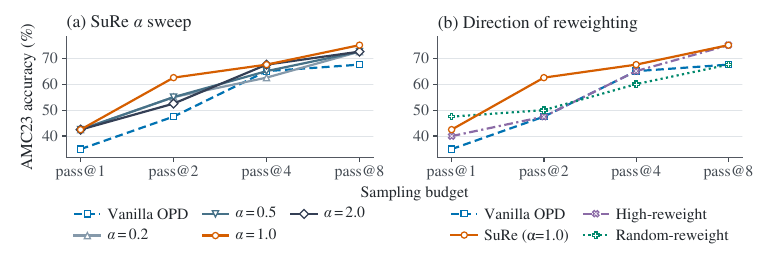}
    \caption{\textbf{Ablations on AMC23 (Qwen3-1.7B-Base).}
    \textbf{(a)} $\alpha$ sweep for SuRe against vanilla OPD.
    \textbf{(b)} Orientation controls comparing SuRe, High-reweight,
    and Random-reweight. SuRe exceeds Random-reweight at $k{=}2,4,8$,
    while Random-reweight scores higher at $k{=}1$. SuRe exceeds
    High-reweight at $k{=}1,2,4$ and ties it at $k{=}8$.}
    \label{fig:ablation-amc23}

\end{figure*}

We conduct two groups of controlled experiments on Qwen3-1.7B-Base
to probe the design choices behind SuRe.
Figure~\ref{fig:ablation-amc23} reports pass@$k$ for
$k{\in}\{1,2,4,8\}$ on AMC23, which exhibits more stable and
discriminative pass@$k$ curves than AIME24/25 at small $k$.

\paragraph{Strength of reweighting ($\alpha$ sweep).}
Figure~\ref{fig:ablation-amc23}(a) sweeps $\alpha\in\{0.2, 0.5, 1.0, 2.0\}$
against vanilla OPD ($\alpha{=}0$). At $k{\in}\{1,2,4\}$,
$\alpha{=}1.0$ attains the highest observed score, including ties;
the scores do not increase monotonically with $\alpha$ at every $k$.
A milder reweighting ($\alpha{=}0.5$) recovers part, but not
all, of the gain, indicating that the effect is not a knife-edge
behaviour around a single setting but a smooth function of how strongly
student-surprised tokens are amplified. Increasing to $\alpha{=}2.0$ weakens
the small-$k$ gain and approaches vanilla OPD on pass@$2$. Thus larger
amplification is not uniformly better; the sweep does not identify a
mechanism beyond this empirical non-monotonicity.

\paragraph{Direction matters.}
At $\alpha=1.0$, Figure~\ref{fig:ablation-amc23}(b) compares SuRe with
\textit{High-reweight}, which favors high-student-probability tokens, and
\textit{Random-reweight}, which permutes weights within each rollout.
The aligned assignment outperforms the opposite assignment in this setting.

\paragraph{How the controls modify OPD.}
All six rows in Table~\ref{tab:matched-controls-main} retain the
sampled-token loss and token-mean denominator in
Eq.~\eqref{eq:method-wsum}; only the detached weights change.
Let $s_t=1-\bar\pi_{S,t}$ and let $\langle\cdot\rangle_{\mathcal T}$
denote the mean over valid response tokens in each micro-batch.
\textit{Vanilla OPD} uses $w_t=1$; \textit{Original SuRe} uses
$w_t=1+\alpha s_t$; \textit{Mean-normalized SuRe} uses
$\tilde w_t=(1+\alpha s_t)/\langle1+\alpha s\rangle_{\mathcal T}$.
\textit{Exact-shuffled} randomly permutes $\tilde w_t$ within
$\mathcal T$; \textit{Exact rank-reversed} assigns the same weights
in reverse surprise order (largest weights to least-surprising tokens).
Both preserve the weight multiset and unit mean.
\textit{Mean-normalized uplift-only} uses
$u_t=1+\alpha s_t\mathbf{1}[\Delta\log p_t>0]$ and
$w_t=u_t/\langle u\rangle_{\mathcal T}$: the surprise increment applies
only to teacher-favored tokens. Other tokens still contribute to the
OPD loss; they are not filtered out.

\begin{center}
\captionof{table}{\textbf{Additional controls on MATH-500 (\%).}
Qwen3-1.7B, seed 42, step 222.}
\label{tab:matched-controls-main}
\small
\setlength{\tabcolsep}{4.2pt}
\begin{tabular}{lcc}
\toprule
\textbf{Training objective} & \textbf{avg@$4$} & \textbf{pass@$4$} \\
\midrule
Vanilla OPD & 66.55 & 80.80 \\
Original SuRe & 67.65 & 80.80 \\
Mean-normalized SuRe & 69.20 & 82.20 \\
Exact-shuffled & 67.95 & 81.00 \\
Exact rank-reversed & 67.00 & 81.40 \\
Mean-normalized uplift-only & 68.40 & 81.40 \\
\bottomrule
\end{tabular}
\end{center}

\paragraph{Matched controls.}
Table~\ref{tab:matched-controls-main} shows that mean-normalized SuRe
preserves the MATH-500 gain, so increased mean weight is not necessary.
Exact-shuffled and rank-reversed weights score numerically lower, but
exact-shuffled still exceeds OPD. Exact surprise assignment is therefore
not isolated from generic non-uniform weighting. Uplift-only improves
over OPD but trails mean-normalized SuRe on avg@$4$. Full results and the
second-seed check are in Appendix~\ref{app:matched-controls}.

\section{Related Work}
\label{sec:related}

\paragraph{Distillation.}
Knowledge distillation (KD) trains smaller students to approximate larger teachers, with sequence-level KD extending this idea to teacher-generated outputs \citep{DBLP:journals/corr/HintonVD15,DBLP:conf/emnlp/KimR16}.
However, such off-policy training suffers from train-inference mismatch because students learn from teacher sequences but decode from their own distributions.
OPD addresses this by training on student rollouts with dense teacher supervision \citep{lu2025onpolicydistillation}.
MiniLLM motivates reverse-KL OPD by its mode-seeking behavior, while GKD broadens the view to mixtures of student- and teacher-generated data \citep{DBLP:conf/iclr/Gu0WH24,DBLP:conf/iclr/AgarwalVZSGGB24}.
Recent work further connects teacher log-ratios to dense KL-constrained RL, making the reward interpretation of token-level distillation explicit \citep{DBLP:journals/corr/abs-2602-12125}.
Our work is complementary: instead of changing the rollout source or divergence, we study how reverse-KL OPD distributes update magnitudes across tokens.

\paragraph{Reasoning Analysis and Optimization.}
Reasoning has advanced through prompting and RL with verifiable rewards (RLVR) \citep{DBLP:conf/nips/Wei0SBIXCLZ22,Shao:2024:GRPO,DBLP:journals/corr/abs-2503-14476}.
Recent analyses show that RLVR gradients concentrate on high-entropy minority tokens and depend strongly on update direction \citep{DBLP:journals/corr/abs-2506-01939,DBLP:journals/corr/abs-2603-22117}.
These findings suggest that not all token positions contribute equally, motivating methods that select or reweight informative tokens rather than matching every position uniformly.
For OPD, prior work studies entropy-aware token selection, probability-based failure filtering, and reward/entropy controls for stabilizing reasoning transfer \citep{DBLP:journals/corr/abs-2603-07079,li2026rethinking,DBLP:journals/corr/abs-2603-11137}.

\section{Conclusion}
\label{sec:conclusion}

We analyzed how the per-token K2 estimator of reverse KL distributes
gradient norms across token positions in on-policy distillation, with the
gradients taken with respect to the student logits. The exact factorization and Qwen3 math diagnostics show a
highly non-uniform distribution, with the largest norms concentrated among
low-student-probability samples that are also enriched in large
teacher--student gaps. SuRe provides one
lightweight test of amplifying this allocation and improves several
metrics in the evaluated setting. Together, these results suggest that
token-level gradient allocation is a useful lens for understanding
sampled-token reverse-KL OPD.

\section{Limitations}
\label{sec:limitations}

First, our current study focuses on sampled token-level reverse-KL OPD,
while alternative design choices, such as full-vocabulary distillation or using Jensen--Shannon divergence as the optimization objective,
have not yet been systematically investigated.
Second, our experiments are primarily conducted on mathematical reasoning datasets,
and the generated reasoning traces have limited length,
which may restrict the generality of our findings to broader domains or substantially longer reasoning processes.
Third, due to computational budget constraints,
we only explore a limited set of model and OPD configurations and do not evaluate our approach on models with larger parameter scales.
Finally, comparisons with closely related OPD reweighting methods under a shared experimental setup remain to be explored.

\bibliography{anthology}

\appendix
\section{Detailed Gradient Derivation}
\label{app:gradient}

This appendix gives the full derivation that supports
Lemma~\ref{lem:opd-grad-norm}, including the choice of the $\ell_1$
norm and a few sanity checks. Throughout, the object of analysis is the
K2 estimator of reverse KL applied at each sampled token: for each fixed
on-policy sampled response token $y_t$, we differentiate the scalar
loss attached to that sampled token and treat the sampled trajectory
itself as fixed during the update. We fix a decoding context
$c_t=(x,y_{<t})$ and write $z\in\mathbb{R}^V$ for the student logits
at position $t$, $p_v\triangleq\pi_S(v\,|\,c_t)$ for the next-token
probability, $p^S_t\triangleq\pi_S(y_t\,|\,c_t)=p_{y_t}$ for the
sampled-token probability, and $\Delta\log p_t=\log\pi_T(y_t\,|\,c_t)-
\log\pi_S(y_t\,|\,c_t)$ for the teacher--student gap of
\eqref{eq:prelim-dlp}. The teacher log-probability
$\log\pi_T(y_t\,|\,c_t)$ is treated as a stop-gradient constant. All
quantities are evaluated at the pre-update student policy. This
appendix does not claim to characterize the full-vocabulary reverse-KL
gradient or the score-function gradient of the sampling distribution.

\subsection{Per-token softmax derivative}
\label{app:gradient-softmax}

\paragraph{Restatement.}
Let $z\in\mathbb{R}^V$ be the student logits and
$\pi_S(v\,|\,c_t)=\exp(z_v)/\sum_{u}\exp(z_u)$ as in
\eqref{eq:prelim-softmax}. For any sampled token $y_t\in\mathcal{V}$,
\begin{equation}
\nabla_z\log\pi_S(y_t\,|\,c_t)
  = e_{y_t} - \pi_S(\cdot\,|\,c_t),
\label{eq:app-grad-logps}
\end{equation}
where $e_{y_t}\in\mathbb{R}^V$ is the one-hot indicator at coordinate
$y_t$. This restates \eqref{eq:prelim-grad-logps}.

\paragraph{Proof.}
Writing the log-probability as
\[
  \log\pi_S(y_t\,|\,c_t)
    = z_{y_t}
    - \log\!\left(\sum_{u\in\mathcal{V}}\exp(z_u)\right),
\]
we differentiate both terms coordinate-wise. For any $v\in\mathcal{V}$,
\begin{align}
\frac{\partial z_{y_t}}{\partial z_v}
  &= \mathbf{1}\{v=y_t\},
\nonumber\\
\frac{\partial}{\partial z_v}
  \log\!\left(\sum_{u}\exp(z_u)\right)
  &= \frac{\exp(z_v)}{\sum_u\exp(z_u)}
\nonumber\\
  &= \pi_S(v\,|\,c_t).
\nonumber
\end{align}
Subtracting these two expressions gives
\begin{align}
\frac{\partial}{\partial z_v}
  \log\pi_S(y_t\,|\,c_t)
  &= \mathbf{1}\{v=y_t\}
\nonumber\\
  &{}- \pi_S(v\,|\,c_t),
\label{eq:app-grad-logps-coord}
\end{align}
which is \eqref{eq:app-grad-logps} stated coordinate-wise.

\subsection{Chain rule for the per-token K2 estimator}
\label{app:gradient-chain}

\paragraph{Restatement.}
For the per-token K2 estimator $L_t^{\mathrm{RKL}}=\tfrac12
(\Delta\log p_t)^2$ from \eqref{eq:prelim-Lt}, its gradient with respect
to the student logits satisfies
\begin{equation}
\nabla_z L_t^{\mathrm{RKL}}
  = -\Delta\log p_t\,\bigl(e_{y_t}-\pi_S(\cdot\,|\,c_t)\bigr),
\label{eq:app-grad-vector}
\end{equation}
which reproduces \eqref{eq:grad-vector}.

\paragraph{Proof.}
Define
\[
  \Delta\log p_t
    = \log\pi_T(y_t\,|\,c_t) - \log\pi_S(y_t\,|\,c_t).
\]
Because $\log\pi_T(y_t\,|\,c_t)$ does not depend on $z$, only the
second term contributes to the gradient, so
\begin{equation}
\nabla_z(\Delta\log p_t)
  = -\nabla_z\log\pi_S(y_t\,|\,c_t).
\label{eq:app-grad-dlp}
\end{equation}
Applying the chain rule to $L_t^{\mathrm{RKL}}=\tfrac12(\Delta\log p_t)^2$,
\begin{align}
\nabla_z L_t^{\mathrm{RKL}}
  &= \Delta\log p_t\cdot\nabla_z(\Delta\log p_t)
\nonumber\\
  &= -\Delta\log p_t\cdot\nabla_z\log\pi_S(y_t\,|\,c_t).
\nonumber
\end{align}
Substituting \eqref{eq:app-grad-logps} for
$\nabla_z\log\pi_S(y_t\,|\,c_t)$ yields \eqref{eq:app-grad-vector}.

\paragraph{Coordinate form.}
The vector identity in \eqref{eq:app-grad-vector} expands into
\begin{equation}
\frac{\partial L_t^{\mathrm{RKL}}}{\partial z_v}
  =
  \begin{cases}
    -\Delta\log p_t\,(1-p^S_t), & v=y_t, \\
    \,\Delta\log p_t\,\pi_S(v\,|\,c_t), & v\neq y_t.
  \end{cases}
\label{eq:app-grad-coord}
\end{equation}
The two branches confirm the descent direction: when $\Delta\log p_t>0$
(teacher endorses the sampled token more than the student), the
$y_t$-coordinate of $-\nabla_z L_t^{\mathrm{RKL}}$ is positive, i.e.\
the update raises the sampled-token logit and lowers all competitor
logits in proportion to $\pi_S(v\,|\,c_t)$; the signs flip when
$\Delta\log p_t<0$.

\subsection{\texorpdfstring{The $\ell_1$ Norm Calculation}{The L1 Norm Calculation}}
\label{app:gradient-l1}

\paragraph{Restatement.}
Under the same sampled token-level reverse-KL setup, the per-token
gradient $\ell_1$ norm satisfies
\begin{equation}
\bigl\|\nabla_z L_t^{\mathrm{RKL}}\bigr\|_1
  = 2\,|\Delta\log p_t|\,(1-p^S_t),
\label{eq:app-grad-l1}
\end{equation}
which is \eqref{eq:opd-grad-l1}.

\paragraph{Proof.}
Using the coordinate form \eqref{eq:app-grad-coord}, split the sum
into the sampled-token term $v=y_t$ and the remaining vocabulary
$v\neq y_t$:
\begin{align}
\bigl\|\nabla_z L_t^{\mathrm{RKL}}\bigr\|_1
  &= \sum_{v\in\mathcal{V}}
       \left|\frac{\partial L_t^{\mathrm{RKL}}}{\partial z_v}\right|
\nonumber\\
  &= |\Delta\log p_t|(1-p^S_t)
\nonumber\\
  &{}+\sum_{v\neq y_t}|\Delta\log p_t|\,\pi_S(v\,|\,c_t).
\label{eq:app-grad-l1-split}
\end{align}
The remaining-coordinate sum simplifies via the simplex identity
$\sum_{v}\pi_S(v\,|\,c_t)=1$:
\begin{equation}
\sum_{v\neq y_t}\pi_S(v\,|\,c_t)
  = 1-p^S_t.
\label{eq:app-simplex}
\end{equation}
Substituting \eqref{eq:app-simplex} into
\eqref{eq:app-grad-l1-split} gives
\begin{align}
\bigl\|\nabla_z L_t^{\mathrm{RKL}}\bigr\|_1
  &= |\Delta\log p_t|(1-p^S_t)
\nonumber\\
  &{}+ |\Delta\log p_t|(1-p^S_t)
\nonumber\\
  &= 2\,|\Delta\log p_t|(1-p^S_t),
\nonumber
\end{align}
which is \eqref{eq:app-grad-l1}.

\paragraph{Geometric reading.}
The two equal halves of \eqref{eq:app-grad-l1} have a clean
geometric meaning. The first half is the absolute magnitude of coordinate
$y_t$ in the gradient with respect to the student logits, and equals
$|\Delta\log p_t|(1-p^S_t)$. The second half is the summed absolute
magnitude across all competitor coordinates, and also equals
$|\Delta\log p_t|(1-p^S_t)$. The two channels point in opposite
directions in logit space but contribute equally to the $\ell_1$ norm,
yielding the factor of $2$ in \eqref{eq:app-grad-l1}.

\subsection{\texorpdfstring{Why We Report the $\ell_1$ Norm}{Why We Report the L1 Norm}}
\label{app:gradient-why-l1}

We measure the per-token update by $\|\nabla_z L_t\|_1$ rather than
$\|\nabla_z L_t\|_2$ for two reasons. (i) The logit-gradient $\ell_1$
norm is the dual sensitivity to $\ell_\infty$-bounded logit perturbations:
\[
\sup_{\|\delta z\|_\infty\leq\epsilon}
\bigl|\langle\nabla_z L_t,\delta z\rangle\bigr|
=\epsilon\|\nabla_z L_t\|_1.
\]
It therefore provides a local logit-space sensitivity measure. For the
first-order softmax response, the total-variation change is bounded by
$\tfrac14\|\delta z\|_1+O(\|\delta z\|_1^2)$. Under the hypothetical
logit-space step $\delta z=-\eta\nabla_zL_t$, this is a local upper-bound
proxy for probability movement, not a characterization of the model's
parameter-space update. (ii) The
$\ell_1$ norm produces a clean multiplicative
factor in $1-\pi_S(y_t\,|\,c_t)$, which is exactly the quantity SuRe
acts on; the $\ell_2$ norm gives an analogous but more algebraically
opaque expression, with $\sqrt{(1-p^S_t)^2+\sum_{v\neq
y_t}\pi_S(v\,|\,c_t)^2}$ replacing the simple $2(1-p^S_t)$ factor.

\subsection{Sanity Check: Confident vs.\ Surprised Tokens}
\label{app:gradient-sanity}

For a confident token at the correct support, $p^S_t\to 1$, the
gradient norm approaches $0$ regardless of $|\Delta\log p_t|$, because
the sampled-token softmax gradient vanishes near saturation.
Quantitatively, both the $y_t$-coordinate magnitude
$|\Delta\log p_t|(1-p^S_t)$ and the summed off-coordinate magnitude
$|\Delta\log p_t|(1-p^S_t)$ vanish jointly. For a surprised token,
$p^S_t\to 0$, the gradient norm approaches $2|\Delta\log p_t|$,
doubling the naive $|\Delta\log p_t|$ estimate one would get from the
$y_t$-coordinate alone; the second factor of $|\Delta\log p_t|$ is the
summed off-coordinate contribution across the rest of the vocabulary.
This is the limit in which the student-probability factor is most visible.

\section{Experimental Details}
\label{app:exp}

\subsection{Pseudo Code of SuRe}
\label{app:sure-algorithm}

To present the SuRe pipeline clearly, we summarize the pseudo code of
SuRe in Algorithm~\ref{alg:sure}. Our implementation is partially
informed by a prior self-distillation training setup, with modifications to
implement the SuRe objective \citep{DBLP:journals/corr/abs-2601-20802}.

\begin{algorithm}[t]
\caption{\textbf{SuRe}: Surprise-aware Reweighted On-Policy Distillation}
\label{alg:sure}
\begin{algorithmic}[1]
\Require Student $\pi_S$ (parameters $\theta$); frozen teacher $\pi_T$; surprise coefficient $\alpha\ge 0$; learning rate $\eta$
\Ensure Updated student parameters $\theta$
\For{step $k = 1, 2, \dots$}
  \State Sample on-policy responses; collect valid token set $\mathcal{T}$
  \ForAll{$t \in \mathcal{T}$}
    \State $\Delta\log p_t \,{:=}\, \log\frac{\pi_T(y_t\,|\,c_t)}{\pi_S(y_t\,|\,c_t)}$
    \State $L_t \,{:=}\, \tfrac{1}{2}(\Delta\log p_t)^2$
    \State $w_t \,{:=}\, 1 + \alpha\bigl(1-\mathrm{sg}(\pi_S(y_t\,|\,c_t))\bigr)$
  \EndFor
  \State $\mathcal{L} \,{:=}\, \dfrac{1}{|\mathcal{T}|}\sum_{t\in\mathcal{T}} w_t\,L_t$ \Comment{\textcolor{gray}{\eqref{eq:method-wsum}}}
  \State $\theta \,{:=}\, \theta{-}\eta\,\nabla_\theta\mathcal{L}$
\EndFor
\end{algorithmic}
\end{algorithm}

\paragraph{Setup.}
Unless otherwise noted, all OPD and SuRe runs use Qwen3-8B as the
frozen teacher, train on the $57$K hard split of DeepMath
\citep{deepmath}, and share the default hyperparameters listed in
Table~\ref{tab:train_hparams}. The student is Qwen3-1.7B-Base for
the main experiments and analysis, and Qwen3-4B-Base for the
cross-scale comparison in Sec.~\ref{sec:experiments-main}. The main runs
use seed $42$. The second-seed check in Table~\ref{tab:second-seed} uses
seed $43$ with all other settings fixed; the corresponding OPD and SuRe
checkpoints are evaluated at step $222$. Each run uses
$4{\times}8 = 32$ H20 GPUs and is trained for two epochs.

\begin{table*}[!ht]
\centering
\caption{\textbf{Training hyperparameters across student scales} for
Vanilla OPD and SuRe. The only difference between the two methods is
the per-token weight $w_t$ in \eqref{eq:method-weight}; the entries
below are shared by the corresponding model-scale runs except for the
memory-related knobs required by the larger student. Rows prefixed by
``Framework'' denote batching limits inherited from the training
implementation. The extra reference-policy KL coefficient is separate from
the K2 distillation objective.}
\label{tab:train_hparams}
\footnotesize
\setlength{\tabcolsep}{5pt}
\renewcommand{\arraystretch}{0.92}
\begin{tabular}{lcc}
\toprule
\textbf{Hyper-parameter}
& \textbf{Qwen3 (8B$\to$1.7B)}
& \textbf{Qwen3 (8B$\to$4B)} \\
\midrule
Student model & Qwen3-1.7B-Base & Qwen3-4B-Base \\
Teacher model & Qwen3-8B & Qwen3-8B \\
Optimizer & AdamW & AdamW \\
Learning rate & $1\times 10^{-6}$ & $1\times 10^{-6}$ \\
LR warmup steps & 10 & 10 \\
Training epochs & 2 & 2 \\
Global batch size & 512 & 512 \\
Framework mini-batch size & 512 & 512 \\
Rollouts per prompt & 1 & 1 \\
Maximum prompt length & 2{,}048 & 2{,}048 \\
Maximum response length & 8{,}192 & 8{,}192 \\
Maximum model length & 12{,}288 & 12{,}288 \\
Framework max token length/GPU & 24{,}576 & 16{,}384 \\
Tensor parallel size & 1 & 1 \\
Generation temperature & 1.0 & 1.0 \\
Top-$p$ (generation) & 1.0 & 1.0 \\
Chunked prefill & enabled & enabled \\
Rollout max batched tokens & 24{,}576 & 16{,}384 \\
Rollout max sequences & 2{,}048 & 512 \\
Rollout GPU memory utilization & 0.92 & 0.80 \\
Actor FSDP size & 8 & 8 \\
Reference FSDP size & 32 & 32 \\
Actor micro-batch/GPU & 2 & 1 \\
Reference log-prob micro-batch/GPU & 4 & 2 \\
Rollout log-prob micro-batch/GPU & 4 & 2 \\
Extra reference-policy KL coefficient & 0.0 & 0.0 \\
Loss aggregation & \texttt{token-mean} & \texttt{token-mean} \\
GPUs & $4{\times}8$ H20 & $4{\times}8$ H20 \\
\bottomrule
\end{tabular}
\end{table*}

\subsection{Prompt Templates}
\label{app:prompt_templates}

\paragraph{Training-time math prompt.}
All math experiments load the $57$K hard split of DeepMath through the
VeRL data pipeline, where every record stores the user message in
the OpenAI chat format. We append a fixed instruction
\texttt{\textbackslash nPlease reason step by step, and put your
final answer within \textbackslash boxed\{\}.} to every user message
that does not already contain \texttt{\textbackslash boxed\{\}}, then
let the tokenizer apply the Qwen3 chat template with
\texttt{enable\_thinking=false}. The resulting training-time prompt
is therefore:

\begin{quote}\small\ttfamily
\noindent$<$|im\_start|$>$user\\
\{problem\}\\
Please reason step by step, and put your final
answer within \textbackslash boxed\{\}.$<$|im\_end|$>$\\
$<$|im\_start|$>$assistant\\
$<$think$>$\\
$<$/think$>$
\end{quote}

\noindent where \texttt{\{problem\}} is the original DeepMath problem
statement. The empty \texttt{$<$think$>$\dots$<$/think$>$} block is
the form that the Qwen3 chat template emits when
\texttt{enable\_thinking=false}; it is part of the prompt and is not
generated by the student. Both the student rollout and the frozen
teacher receive exactly the same prompt, so the teacher--student
gap $\Delta\log p_t$ used in
\eqref{eq:prelim-dlp} is computed under aligned contexts.

\paragraph{Evaluation-time math prompt.}
Evaluation prompts use the same instruction text appended to the
problem statement, so that DeepMath-trained students see a prompt
distribution at test time that matches the training distribution:

\begin{quote}\small\ttfamily
\noindent\{problem\}\\
Please reason step by step, and put your final
answer within \textbackslash boxed\{\}.
\end{quote}

\noindent For the in-domain math benchmarks (AIME2024, AIME2025, AMC23,
and MATH-500), responses are parsed by extracting the last
\texttt{\textbackslash boxed\{\dots\}} expression and compared against
ground-truth answers via exact match or the corresponding benchmark verifier.

\section{Additional Experimental Results}
\label{app:add-results}

\subsection{Full Out-of-Domain Results}
\label{app:full-ood}

Table~\ref{tab:full-ood-passk} reports the full pass@$k$ results on
all OOD benchmarks used in the main text. The main paper reports
pass@1 for compactness, while this appendix includes pass@1, pass@5,
and pass@10 to expose the best-of-$k$ behavior of each method. For
IFEval we use prompt-level strict accuracy; for MMLU-Pro we use exact
match.

\begin{table*}[t]
\centering
\caption{\textbf{Full OOD pass@$k$ results (\%).}
CRUX, IFEval, and MMLU-Pro are the OOD benchmarks reported in the main
text. IFEval uses prompt-level strict accuracy and MMLU-Pro uses exact
match.}
\label{tab:full-ood-passk}
\scriptsize
\setlength{\tabcolsep}{3.2pt}
\renewcommand{\arraystretch}{0.92}
\resizebox{\textwidth}{!}{
\begin{tabular}{l|ccc|ccc|ccc}
\toprule
& \multicolumn{3}{c|}{\textbf{CRUX}} & \multicolumn{3}{c|}{\textbf{IFEval}} & \multicolumn{3}{c}{\textbf{MMLU-Pro}} \\
\cmidrule{2-4} \cmidrule{5-7} \cmidrule{8-10}
\textbf{Method} & \textbf{@1} & \textbf{@5} & \textbf{@10} & \textbf{@1} & \textbf{@5} & \textbf{@10} & \textbf{@1} & \textbf{@5} & \textbf{@10} \\
\midrule
\multicolumn{10}{c}{\textit{\textbf{Qwen3-1.7B-Base}}} \\
\midrule
Base & 4.75 & 21.75 & 33.12 & 22.92 & 47.50 & 57.86 & 27.07 & 62.94 & 78.17 \\
Vanilla OPD & 35.38 & 63.00 & 70.62 & 27.73 & 44.18 & 53.97 & 26.44 & 57.32 & 69.91 \\
SuRe ($\alpha{=}0.2$) & 34.62 & 63.25 & 71.12 & 26.80 & 44.92 & 53.05 & 26.52 & 57.37 & 69.98 \\
SuRe ($\alpha{=}0.5$) & 34.75 & 60.88 & 68.50 & 27.17 & 44.92 & 52.68 & 26.86 & 57.31 & 69.56 \\
SuRe ($\alpha{=}1.0$) & 34.88 & 63.62 & 70.25 & 28.10 & 43.25 & 52.31 & 25.90 & 57.67 & 69.92 \\
SuRe ($\alpha{=}2.0$) & 34.00 & 62.25 & 69.38 & 29.57 & 45.47 & 51.76 & 28.09 & 58.30 & 70.35 \\
High-reweight & 32.00 & 61.88 & 69.62 & 29.21 & 46.03 & 51.57 & 27.29 & 57.55 & 69.92 \\
Random-reweight & 33.75 & 60.50 & 69.50 & 29.76 & 45.10 & 51.57 & 27.34 & 57.80 & 70.00 \\
\specialrule{1pt}{0.4ex}{0.4ex}
\multicolumn{10}{c}{\textit{\textbf{Qwen3-4B-Base}}} \\
\midrule
Base & 23.00 & 56.50 & 67.25 & 31.79 & 56.56 & 66.17 & 38.71 & 72.31 & 83.02 \\
Vanilla OPD & 59.38 & 80.38 & 84.88 & 36.23 & 56.93 & 63.22 & 42.39 & 73.51 & 82.01 \\
SuRe ($\alpha{=}1.0$) & 59.25 & 80.12 & 84.75 & 35.30 & 55.82 & 64.33 & 42.70 & 72.37 & 81.75 \\
\bottomrule
\end{tabular}}
\end{table*}

\subsection{Training Dynamics on Qwen3-4B-Base}
\label{app:train-dynamics-4b}

\begin{figure*}[!ht]
    \centering
    \begin{subfigure}[t]{0.33\textwidth}
        \centering
        \includegraphics[width=\linewidth]{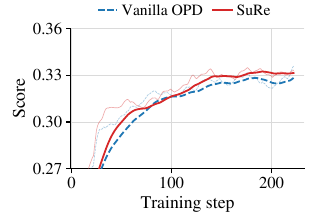}
        \caption{Mean rollout score}
        \label{fig:train-dyn-score-4b}
    \end{subfigure}%
    \begin{subfigure}[t]{0.33\textwidth}
        \centering
        \includegraphics[width=\linewidth]{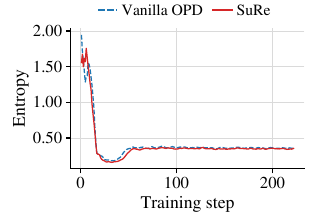}
        \caption{Actor entropy}
        \label{fig:train-dyn-ent-4b}
    \end{subfigure}%
    \begin{subfigure}[t]{0.33\textwidth}
        \centering
        \includegraphics[width=\linewidth]{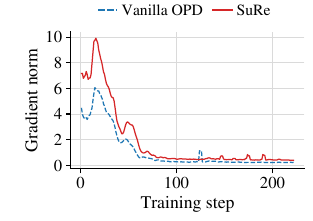}
        \caption{Gradient norm}
        \label{fig:train-dyn-grad-4b}
    \end{subfigure}
    \caption{\textbf{Training dynamics on Qwen3-4B-Base.}
    Vanilla OPD vs.\ SuRe at $\alpha{=}1.0$. SuRe maintains comparable
    actor entropy while changing the score and gradient-norm dynamics,
    providing an additional cross-scale view of the optimization
    behavior.}
    \label{fig:train-dynamics-4b}
\end{figure*}

Figure~\ref{fig:train-dynamics-4b} provides the same training-dynamics
view as Figure~\ref{fig:train-dynamics}, but for Qwen3-4B-Base. The
curves complement the 1.7B analysis by showing that the qualitative
optimization behavior remains similar at the larger student scale.

\subsection{Token-level Case Study}
\label{app:case-study}

\paragraph{Takeaway.}
This case study is intended to answer a narrow descriptive question:
\emph{where} do the final OPD checkpoint and the Base initialization
differ on a concrete sampled solution? In this correct MATH-500 rollout,
most response tokens remain close across the two checkpoints, while a
small number of local decision points differ sharply.

\begin{table*}[!ht]
\centering
\caption{\textbf{A compact view of the case.} The example is a correct
OPD rollout. We keep only the information needed to interpret the
subsequent token table, rather than listing every logged field.}
\label{tab:case-study-card}
\small
\setlength{\tabcolsep}{5pt}
\renewcommand{\arraystretch}{1.08}
\begin{tabular}{p{0.18\textwidth}p{0.76\textwidth}}
\toprule
\textbf{Item} & \textbf{Content} \\
\midrule
Task & MATH-500 example asking for primes $p$ such that $8x\equiv 1\pmod p$ has no solution. \\
Config & Qwen3-1.7B OPD student at step $222$; Qwen3-1.7B-Base as base; Qwen3-8B as teacher; temperature $1.0$, top-$p=1.0$, max length $8192$. \\
Prompt & ``Determine the sum of all such $p$. Please reason step by step, and put your final answer within \texttt{\textbackslash boxed\{\}}.'' \\
Response sketch & The student argues that the congruence is solvable iff $\gcd(8,p)=1$; only the prime $p=2$ divides $8$; therefore the answer is $\boxed{2}$. \\
Global pattern & $376$ response tokens; mean JS$(\pi_{\mathrm{OPD}},\pi_{\mathrm{base}})=0.052$; only $16.8\%$ of tokens have JS$>0.10$. \\
\bottomrule
\end{tabular}
\end{table*}

\paragraph{Token highlights.}
Table~\ref{tab:case-study-tokens} gives the same example in the most
compact token-level form: the token location, the log-probability
difference between the final OPD checkpoint and the Base initialization,
and the plain-language reading. Positive $\Delta_{\mathrm{OPD-Base},t}$ means the final
OPD checkpoint assigns higher probability to the sampled token than Base;
negative $\Delta_{\mathrm{OPD-Base},t}$ means it assigns lower probability.

\begin{table*}[!ht]
\centering
\caption{\textbf{Four readable token-level changes from the case.}
The point is not that every large change is a mathematical token, but
that the endpoint difference is localized: a few reasoning, formatting,
and transition positions differ strongly instead of the whole response
moving uniformly.}
\label{tab:case-study-tokens}
\small
\setlength{\tabcolsep}{4pt}
\renewcommand{\arraystretch}{1.12}
\begin{tabular}{p{0.18\textwidth}p{0.20\textwidth}p{0.17\textwidth}p{0.37\textwidth}}
\toprule
\textbf{Where} & \textbf{Sampled token} & \textbf{OPD vs. base} & \textbf{Interpretation} \\
\midrule
After ``analyze the given congruence'' & newline after ``:'' & rank $5\to1$, $\Delta_{\mathrm{OPD-Base},t}=+2.95$ & The final OPD checkpoint ranks the transition into displayed-equation format more highly than Base. \\
At the key concept ``invertible modulo $p$'' & emphasis marker before ``invertible'' & rank $6\to1$, $\Delta_{\mathrm{OPD-Base},t}=+6.82$ & The largest endpoint difference occurs where the solution introduces the modular-inverse idea. \\
At the restatement ``which primes $p$ ...'' & the transition token ``the'' & rank $5\to1$, $\Delta_{\mathrm{OPD-Base},t}=+2.90$ & Some large endpoint differences occur on discourse or fluency tokens, not only mathematical symbols. \\
At an intermediate mention of the answer & token ``2'' & rank $2\to2$, $\Delta_{\mathrm{OPD-Base},t}=-2.70$ & The final OPD checkpoint assigns this token lower probability than Base; the final answer remains correct, but the local confidence profile differs. \\
\bottomrule
\end{tabular}
\end{table*}

\noindent
In short, this case provides a qualitative endpoint comparison:
\textbf{the final OPD checkpoint differs from the Base initialization
most strongly at a small number of local token positions rather than
uniformly across the response.}

\subsection{Evaluation}
\label{app:evaluation}

\paragraph{Evaluation.}
We evaluate math reasoning with a standalone evaluation pipeline.
Generation uses temperature $0.7$, top-$p$ $0.9$, maximum generation
length $31{,}744$, bfloat16 inference, and the prompt template in
Sec.~\ref{app:prompt_templates}. Consistent with the main text, the
appendix math results only include AIME24 (30 problems), AIME25 (30
problems), AMC23 (40 problems), and MATH-500 (500 problems). For each
problem in AIME24, AIME25, and AMC23, we generate $32$ rollouts,
but compute the reported avg@$8$ and pass@$8$ using only the first
eight samples in generation order. avg@$8$ is the mean correctness across these eight
samples, averaged over problems; pass@$8$ is the fraction of problems
with at least one correct answer among these eight samples. The remaining
24 samples are not used for these @$8$ metrics. For MATH-500, avg@$4$
and pass@$4$ use all four samples with the same definitions.
All reported results are computed using a heuristic grader.

\subsection{Comparison with On-Policy RL (GRPO)}
\label{app:grpo}

The main paper analyzes OPD trained with the per-token K2 estimator and uses SuRe as an intervention,
with KD, SeqKD, and Vanilla OPD as baselines. For completeness, we additionally
compare SuRe against GRPO~\citep{Shao:2024:GRPO}, a reward-based
on-policy RL method, on Qwen3-1.7B-Base. We adopt this scale because
the ablations and orientation controls in
Sec.~\ref{sec:experiments-controls} are also conducted on
Qwen3-1.7B-Base. We stress that on-policy RL methods such as GRPO and
on-policy distillation are \emph{complementary} rather than mutually
exclusive: GRPO learns from a verifiable reward signal, whereas SuRe
learns from a stronger teacher's token-level distribution, so the two
signals can in principle be combined. We therefore did not place the
GRPO comparison in the main table, and instead report it here as an
additional reference point.

\paragraph{Setup.}
GRPO is trained with a verifiable answer-matching reward on the same
DeepMath hard split, using the same teacher prompt template, the same
$32{\times}$H20 setup, and the same evaluation protocol as our SuRe
runs (Sec.~\ref{app:evaluation}). During evaluation, both methods are
decoded with the same temperature ($0.7$), top-$p$ ($0.9$), and number
of sampled solutions per problem ($N{=}32$ for AIME24/AIME25/AMC23,
$N{=}4$ for MATH-500). For both methods, avg@$8$/pass@$8$ use only
the first eight samples in generation order, and avg@$4$/pass@$4$
use all four MATH-500 samples.

\paragraph{Baseline configurations.}
For reproducibility, Table~\ref{tab:baseline_hparams} lists the
training hyperparameters used for the KD, SeqKD, and GRPO baselines on
Qwen3-1.7B-Base. KD and SeqKD are run on $8$ H20 GPUs. Entries that
match the OPD/SuRe defaults in Table~\ref{tab:train_hparams}
(optimizer, learning rate, warmup, epochs, batch sizes, sequence
lengths, and FSDP/parallelism) are omitted to avoid duplication; only
the method-specific knobs are shown.

\begin{table*}[!t]
\centering
\caption{\textbf{Baseline-specific hyperparameters} for KD, SeqKD, and
GRPO on Qwen3-1.7B-Base. Entries shared with OPD/SuRe (see
Table~\ref{tab:train_hparams}) are omitted.}
\label{tab:baseline_hparams}
\footnotesize
\setlength{\tabcolsep}{4pt}
\renewcommand{\arraystretch}{0.92}
\resizebox{\textwidth}{!}{
\begin{tabular}{lccc}
\toprule
\textbf{Hyper-parameter} & \textbf{KD} & \textbf{SeqKD} & \textbf{GRPO} \\
\midrule
Data source & DeepMath hard & Teacher rollouts & DeepMath hard \\
Teacher model & Qwen3-8B & Qwen3-8B & -- \\
Supervision & Token-level $\pi_T$ & Sequence sampled from $\pi_T$ & Verifiable answer reward \\
Loss & Forward KL on teacher tokens & SFT cross-entropy on teacher seq. & GRPO clipped policy gradient \\
Rollout sampling & off-policy (teacher) & off-policy (teacher) & on-policy (student) \\
Generation temperature & -- & 1.0 & 1.0 \\
Top-$p$ (generation) & -- & 1.0 & 1.0 \\
Rollouts per prompt & -- & 1 & 8 \\
Group size $G$ & -- & -- & 8 \\
Extra reference-policy KL coefficient & 0.0 & 0.0 & 0.001 \\
Clip ratio $\epsilon$ & -- & -- & 0.2 \\
Reward & -- & -- & $\{0,1\}$ exact match on \texttt{\textbackslash boxed\{\}} \\
Loss aggregation & \texttt{token-mean} & \texttt{token-mean} & \texttt{token-mean} \\
\bottomrule
\end{tabular}}
\end{table*}

\begin{table*}[!t]
    \centering
    \caption{\textbf{Comparison with on-policy RL on Qwen3-1.7B-Base (\%).}
    GRPO is a reward-based RL baseline trained with verifiable answer
    matching; SuRe is purely a distillation objective with no reward
    signal. AIME24, AIME25, and AMC23 compute avg@$8$/pass@$8$ from
    only the first $8$ of $32$ rollouts per problem, in generation order;
    MATH-500 computes avg@$4$/pass@$4$ from all $4$ rollouts.
    \textbf{Bold} marks the better entry within each column.}
    \label{tab:grpo}
    \small
    \setlength{\tabcolsep}{5pt}
    \renewcommand{\arraystretch}{1.12}
    \resizebox{\textwidth}{!}{
    \begin{tabular}{l|cc|cc|cc|cc}
    \toprule
        & \multicolumn{2}{c|}{\textbf{AIME24}}        & \multicolumn{2}{c|}{\textbf{AIME25}}
        & \multicolumn{2}{c|}{\textbf{AMC23}}
        & \multicolumn{2}{c}{\textbf{MATH-500}} \\
        \cmidrule{2-3} \cmidrule{4-5} \cmidrule{6-7} \cmidrule{8-9}
        \textbf{Method}
        & avg@$8$ & pass@$8$
        & avg@$8$ & pass@$8$
        & avg@$8$ & pass@$8$
        & avg@$4$ & pass@$4$ \\
        \midrule
        GRPO
        & 8.33  & \textbf{23.33}
        & 4.17  & \textbf{20.00}
        & 41.56 & 70.00
        & 65.80 & 79.00 \\
        SuRe ($\alpha{=}1.0$)
        & \textbf{9.58}  & \textbf{23.33}
        & \textbf{7.08}  & 16.67
        & \textbf{43.12} & \textbf{75.00}
        & \textbf{67.65} & \textbf{80.80} \\
    \bottomrule
    \end{tabular}}
\end{table*}

\paragraph{Findings.}
Table~\ref{tab:grpo} shows that SuRe matches or exceeds GRPO on every
benchmark in avg@$k$, with consistent improvements of roughly
$1$--$3$pp across AIME24, AIME25, AMC23, and MATH-500. The pass@$k$
picture is mixed: SuRe ties or wins on three of four benchmarks but trails
GRPO on AIME25. Because the methods use different supervision and we do
not report a significance test for this comparison, Table~\ref{tab:grpo}
is an additional reference point rather than evidence that the objectives
are equivalent.

\subsection{Matched Weighting Controls}
\label{app:matched-controls}

Mean normalization divides the detached weights by their mean over valid
response tokens within each micro-batch. Exact-shuffled permutes those
normalized weights within the same token set. Exact rank reversal
preserves the realized weight multiset but reverses its surprise-rank
assignment. Mean-normalized uplift-only applies the surprise increment
only when the signed teacher--student log-probability gap is positive and
then uses the same normalization.

\begin{table*}[!t]
\centering
\caption{\textbf{Matched weighting controls on Qwen3-1.7B-Base (\%).}
AIME24, AIME25, and AMC23 report avg@$8$/pass@$8$; MATH-500 reports
avg@$4$/pass@$4$. Rank-reversed was evaluated only on MATH-500.}
\label{tab:matched-controls-full}
\small
\setlength{\tabcolsep}{4.2pt}
\renewcommand{\arraystretch}{1.05}
\resizebox{\textwidth}{!}{
\begin{tabular}{l|cc|cc|cc|cc}
\toprule
& \multicolumn{2}{c|}{\textbf{AIME24}}
& \multicolumn{2}{c|}{\textbf{AIME25}}
& \multicolumn{2}{c|}{\textbf{AMC23}}
& \multicolumn{2}{c}{\textbf{MATH-500}} \\
\cmidrule{2-3} \cmidrule{4-5} \cmidrule{6-7} \cmidrule{8-9}
\textbf{Training objective}
& avg@$8$ & pass@$8$
& avg@$8$ & pass@$8$
& avg@$8$ & pass@$8$
& avg@$4$ & pass@$4$ \\
\midrule
Vanilla OPD
& 9.17 & 16.67 & 5.83 & 16.67 & 39.38 & 67.50 & 66.55 & 80.80 \\
Mean-normalized SuRe
& 10.00 & 20.00 & 4.58 & 10.00 & 41.88 & 72.50 & 69.20 & 82.20 \\
Exact-shuffled
& 10.00 & 23.33 & 5.83 & 23.33 & 38.44 & 70.00 & 67.95 & 81.00 \\
Exact rank-reversed
& -- & -- & -- & -- & -- & -- & 67.00 & 81.40 \\
Mean-normalized uplift-only
& 9.58 & 26.67 & 4.17 & 13.33 & 39.38 & 72.50 & 68.40 & 81.40 \\
\bottomrule
\end{tabular}}
\end{table*}

\begin{table}[!t]
\centering
\caption{\textbf{Second-seed check on MATH-500 avg@$4$ (\%).}
Original SuRe uses the submitted, unnormalized weighting rule.}
\label{tab:second-seed}
\small
\setlength{\tabcolsep}{7pt}
\begin{tabular}{ccc}
\toprule
\textbf{Seed} & \textbf{Vanilla OPD} & \textbf{Original SuRe} \\
\midrule
42 & 66.55 & 67.65 \\
43 & 66.40 & 67.50 \\
\bottomrule
\end{tabular}
\end{table}

\end{document}